\documentclass[conference]{IEEEtran}
\IEEEoverridecommandlockouts
\usepackage{cite}
\usepackage{amsmath,amssymb,amsfonts}
\usepackage{textcomp}
\usepackage{xcolor}

\usepackage{booktabs} 
\usepackage{mathtools}
\usepackage{subcaption}

\usepackage{algorithm}
\usepackage[noend]{algpseudocode}
\usepackage[T1]{fontenc}
\usepackage{import}
\usepackage{graphicx}
\graphicspath{ {./images/} }

\def\BibTeX{{\rm B\kern-.05em{\sc i\kern-.025em b}\kern-.08em
    T\kern-.1667em\lower.7ex\hbox{E}\kern-.125emX}}
\begin{document}

\title{Frosting Weights for Better Continual Training}

\author{\IEEEauthorblockN{Xiaofeng Zhu}\thanks{Xiaofeng Zhu and Feng Liu contributed equally to this work. Xiaofeng Zhu is the corresponding author.}
\IEEEauthorblockA{\textit{Department of CS} \\
\textit{Northwestern University}\\
Evanston, USA \\
xiaofengzhu2013@u.northwestern.edu}
\and
\IEEEauthorblockN{Feng Liu}
\IEEEauthorblockA{\textit{Department of CEECS} \\
\textit{Florida Atlantic University}\\
Boca Raton, USA \\
fliu2016@fau.edu}
\and
\IEEEauthorblockN{Goce Trajcevski}
\IEEEauthorblockA{\textit{Department of ECpE} \\
\textit{Iowa State University}\\
Ames, USA \\
gocet25@iastate.edu}
\and
\IEEEauthorblockN{Dingding Wang}
\IEEEauthorblockA{\textit{Department of CEECS} \\
\textit{Florida Atlantic University}\\
Boca Raton, USA \\
wangd@fau.edu
}
}
\maketitle

\begin{abstract}
Training a neural network model can be a lifelong learning process and is a computationally intensive one. A severe adverse effect that may occur in deep neural network models is that they can suffer from catastrophic forgetting during retraining on new data. To avoid such disruptions in the continuous learning, one appealing property is the additive nature of ensemble models. In this paper, we propose two generic ensemble approaches, gradient boosting and meta-learning, to solve the catastrophic forgetting problem in tuning pre-trained neural network models.

\end{abstract}


\maketitle

\section{Introduction}

With stationary training resources and various advanced neural network structures, deep learning models have exceeded human performance in many areas. However, a well-known limitation of deep learning models is the so-called ``catastrophic forgetting.'' That is, while acquiring new knowledge, the models may forget the knowledge learned in the past. In other words, after tuning a pre-trained model with new data, the model can have a poor performance on patterns learned from old data.

In this paper, we investigate the possibility of a model to generalize from old patterns and new patterns. Towards formalizing the setting of our study, we assume that we have a model with a generic loss function, and it has been well-trained on a big dataset. Our aim is that, upon presenting a new dataset, 
the model can be adjusted so that it can handle the new dataset well -- while simultaneously  not forgetting  the patterns learned from the old dataset. In many real-world systems, new and old data are commonly stored in a buffered manner, where the model always has access to a roughly fixed size of training data. This fact is at the core of what differentiates our model retraining from traditional continual learning with bounded computational complexity. In our study, the old and the new datasets share the same prediction space. For instance, the number of classes in the two datasets are the same, but the background and object angles may differ.

The issue of catastrophic forgetting in neural networks had been addressed in the literature (cf.~\cite{french1999catastrophic, kirkpatrick2017overcoming}). Once again, the source of the problem is that when neural networks are trained continually on datasets with different feature distributions and/or trained for different objectives, the knowledge learned in an old training session can be lost in a new training session ~\cite{yosinski2014transferable}.

We note that our study aligns closely to continual/incremental learning and multi-task/sequential learning. Continual learning focuses on tuning a model with continual new data without forgetting old knowledge~\cite{albesano2006adaptation, silver2013lifelong, ruvolo2013ella, li2017learning, DBLP:journals/corr/abs-1802-07569}. 
Compared to continual learning, multi-task learning focuses on more dissimilar tasks~\cite{Ruder2017, zhang2017, henderson2017benchmark, rusu2016progressive}, e.g., incremental class learning~\cite{roy2018tree, rebuffi2017icarl, kemker2017fearnet, Ganegedara2017, li2018}, where different tasks have different classes. Moreover, different tasks usually only have access to new data~\cite{aljundi2018selfless} or just a small portion of old data~\cite{lopez2017gradient}. From the training objective aspect, multi-task learning solves task-specific problems: parameter space shifting using priors~\cite{lee2017overcoming, nguyen2018variational, swaroop2019improving}, path traversal among different tasks~\cite{lopez2017gradient, Fernando2017}, parameter sharing in different tasks~\cite{Elliot2018}, and hard attentions~\cite{serra2018overcoming} or masks~\cite{mallya2018packnet} for switching tasks. Their contributions address challenges in different experimental settings~\cite{Maltoni2018}, such as how data in different tasks are distributed, how tasks are ordered~\cite{Meyerson2018, standley2019tasks}, and how performance should be evaluated among a sequence of tasks.

Our study may be perceived as a special case of continual learning -- however, we note that our primary focus and data setting differ from continual learning and, especially, sequential learning. Our retraining models aim to have good performance on both new and old data. In continual learning and sequential learning, a model can only access new data. However, when tuning a pre-trained model in real-world systems, the model should access some old data to guarantee adequate retraining. It may not be feasible in the long run to train on the union of all old data and new data. In our study, old data and new data are stored in a buffered manner. Given limited computation resources that can process a fixed capacity of data, we feed the same full capacity of training data during training or retaining. In every future retraining process, a trained model is tuned on a fixed-size data set, which is composed of the new dataset and a subset of the old dataset. We boost the performance of a trained model by gradient boosting and by learning a meta-network. The two methods are generic and can be applied to any neural network models and loss functions.

Our major contributions are that (1) we propose an additive ensemble model that can tune a pre-trained model based on its performance gap, and (2) we further propose a meta-network model that can balance new patterns and old patterns and eventually keep the number of parameters unchanged.

\section{Related Work}
\label{Related Work}
We now detail the major directions of continual learning: regularization, ensemble methods, and memory consolidation.

\subsection{Regularization}
Regularization is the most common way of solving catastrophic forgetting via approaches that measure the importance of weights of a trained model. The elastic weight consolidation (EWC) algorithm proposed in~\cite{kirkpatrick2017overcoming} uses the Fisher matrix to represent weight importance. A regularization term is added to the original loss function to constrain important weights to stay close to their old values when retraining on a new session/task. Despite the usefulness of the Fisher matrix, the diagonal Fisher matrix assumption may not hold; therefore, approximations of the Fisher Matrix~\cite{tu2016ranking, liu2018rotate, Schwarz2018, zeno2018task} and additional KL-divergence constraints~\cite{chaudhry2018riemannian} were applied in later studies. Besides the Fisher matrix, gradient magnitude~\cite{aljundi2018memory}, unsupervised self-organizing maps~\cite{kohonen1990self}, and the Hessian matrix~\cite{zenke2017continual} are common techniques for measuring weight importance. In particular, the memory aware synapses (MAS) model~\cite{aljundi2018memory} has become a new widely-used benchmark after EWC. MAS uses the gradient magnitude of network outputs with regard to weights to measure weight importance. In contrast, the Fisher matrix is based on loss and weights in EWC. When outputs are multi-dimensional, as in classification tasks, MAS calculates the squared l2 norm of outputs. An improvement over EWC and MAS by adding an additional regularization term on neurons was presented in~\cite{aljundi2018selfless}, proposing that sparsity at the neuron level was also important. We show in our experiments that our ensemble models perform better than this regularization strategy.

\subsection{Ensemble Methods}
Intuitively, ensemble models try to solve catastrophic forgetting by training multiple sub-networks~\cite{polikar2001learn++, Dai2017, hinton2015distilling, chen2017learning, wang2017model, YuSlimmable, Luong2016}. Ren et al.~\cite{Ren2017} and Juefei-Xu et al.~\cite{juefei-xu2018pnn} made breakthroughs by proposing dynamic ways of constructing an ensemble model by tuning a sub-network of a pre-trained model or expanding a pre-trained model. The general limitation is that the memory usage scales up with the number of training sessions. Related meta-learning works generally focus on learning network architectures in order to adjust a pre-trained model to be robust on different datasets and tasks~\cite{yoon2018lifelong, lin2013network, Rusu2016, xu2018reinforced, javed2019meta, juefei-xu2018pnn}. For instance, knowledge distillation~\cite{hinton2015distilling} tries to distill knowledge learned from a large (teacher) network to a small (student) network. The goal is to make the teacher network and the student network behave as similarly as possible. However, our goal is to learn a more generalized network continually, being able to fix errors in the old trained model by acquiring new data.

\subsection{Memory Consolidation}
Memory consolidation solves catastrophic forgetting from the data aspect. Memory consolidation models~\cite{kemker2017fearnet, lopez2017gradient, Nitin2017, chaudhry2018efficient, chaudhry2019continual, Mehta2018, Sodhani2019} learn to store the patterns/memories of different training sessions by strategically sampling a subset of the old data. Generative Adversarial Networks (GANs) are suitable in nature for learning networks~\cite{shin2017continual, Seff2017, Rios2018, Thanh-Tung2018, liang2019generative, Aljundi2019, Subakan2019} in an incremental way. A generator trained on old data can be used in a new training session for replaying old learning experiences. Our study solves a more generic issue from the modeling aspect, and memory consolidation can be applied to our models for providing a reasonable data buffer.

\begin{figure*}[pt]
\begin{subfigure}[b]{0.5\textwidth}
  \includegraphics[width = \textwidth]{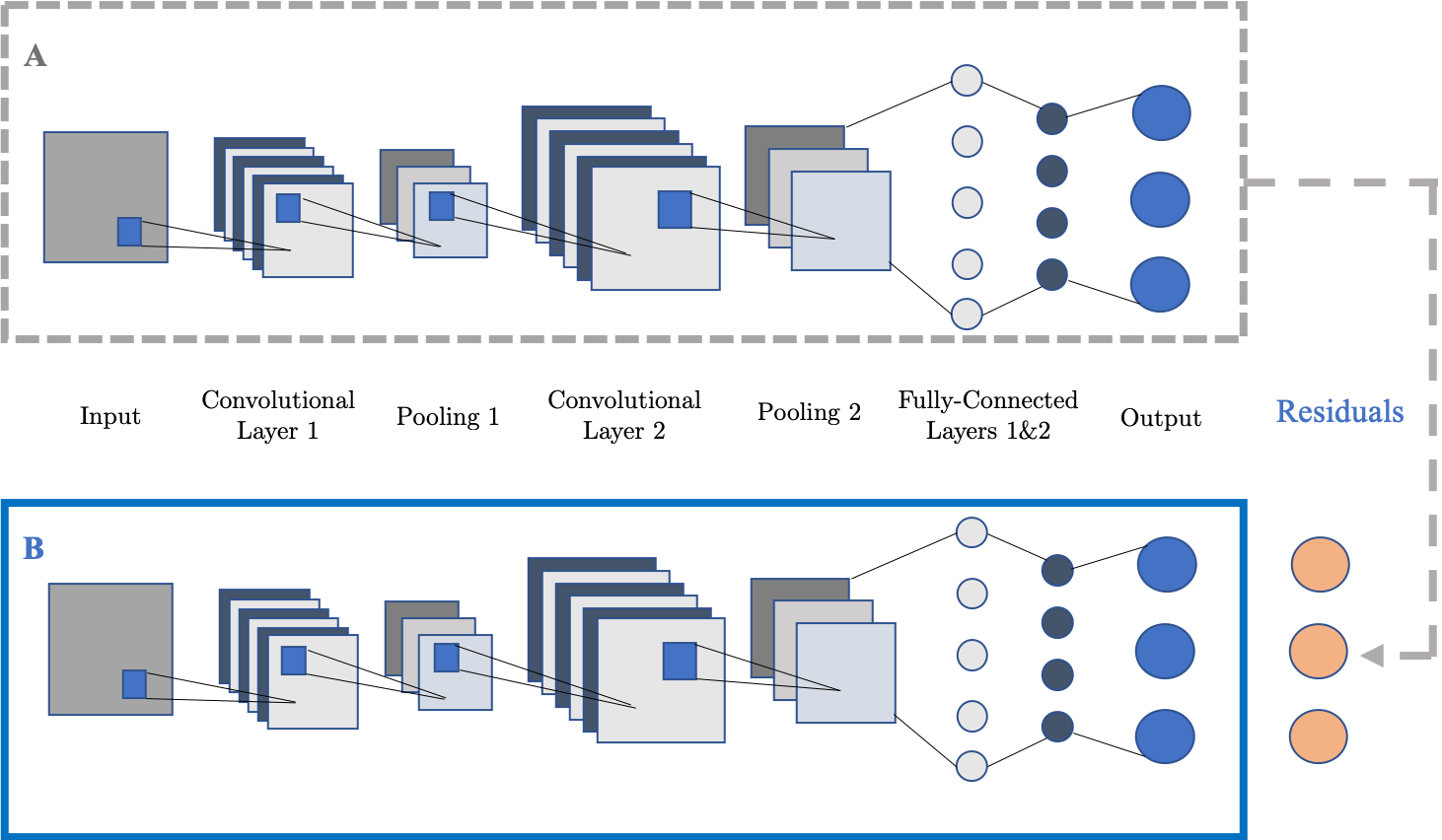}
  \caption{BoostNet}
  \label{fig:BoostNet}
\end{subfigure}\hspace{0.1\textwidth}%
\begin{subfigure}[b]{0.5\textwidth}
  \includegraphics[width =0.86\textwidth]{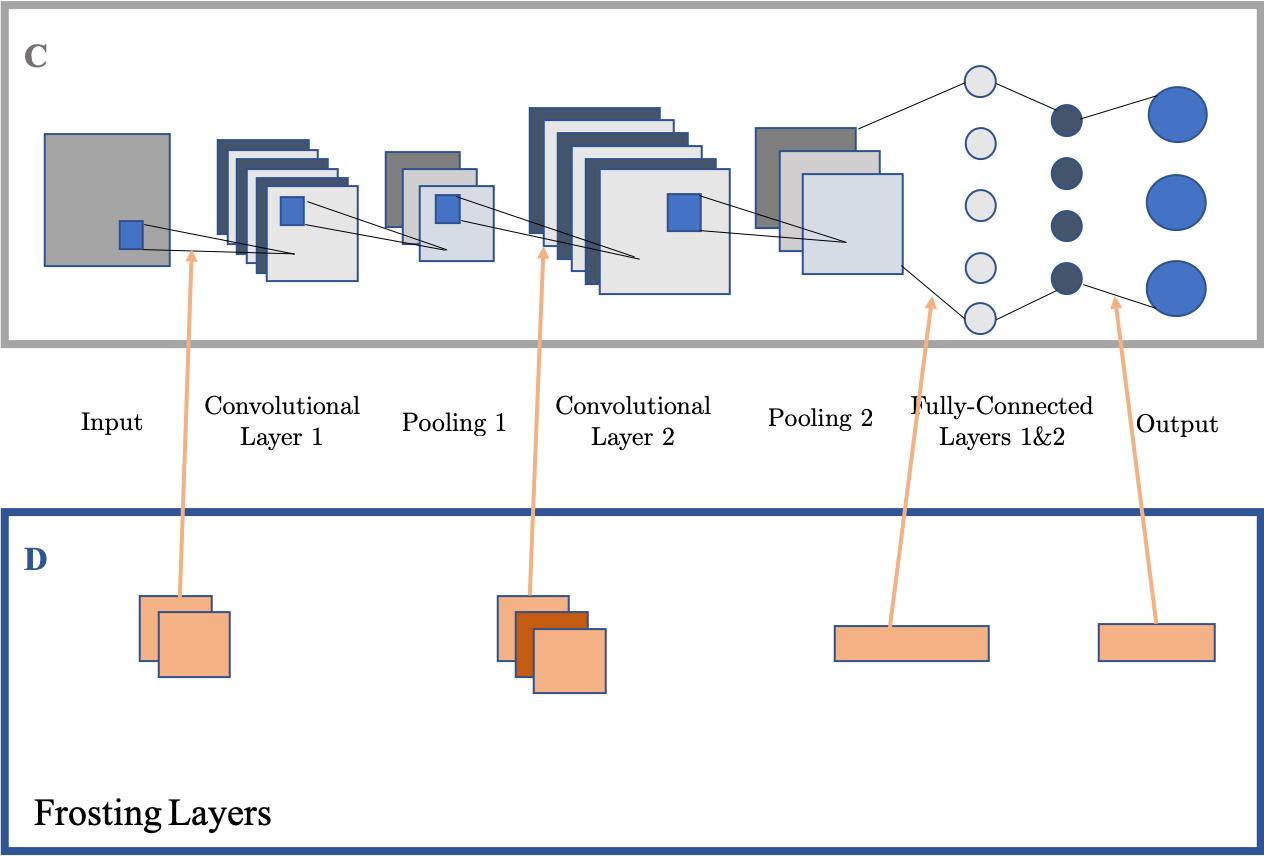}
  \caption{FrostNet}
  \label{fig:FrostNet}
\end{subfigure}
\end{figure*}

\section{Proposed Models}
\label{Proposed Models}

In this section, we present two ways of ``frosting'' weights for better retraining. One is based on classical gradient boosting, the other is based on neural network meta-learning.

\subsection{Frosting Network Weights by One-step Gradient Boosting}
\label{Frosting Network Weights by One-step Gradient Boosting}

We train our proposed network BoostNet, which can be simpler or the same as the trained network. We firstly apply a retraining dataset, composed of new training data and a subset of the old training data, to the trained model and calculate the model outputs and ``residuals''~\cite{friedman2001greedy}. Let $L$ denote the softmax cross-entropy loss function, $o_i$ denote the output of the $i^{th}$ neuron before the softmax layer, $p_i$ denote the softmax probability of the $i^{th}$ neuron, and let $y_i$ denote the one-hot label. The residuals are calculated as $\frac{\partial L}{\partial o_i} = y_i - p_i$, measuring how much the trained model needs to improve in order to perform well on the retraining dataset. It is very likely that the residual values are close to 0 for the old training data, since the loss was originally minimized, and the residual values for the new training data are relatively larger. The weights of the trained model stay unchanged. Figure~\ref{fig:BoostNet} illustrates this training procedure. The BoostNet takes the residuals as the training targets (``labels''). The loss function for the simpler model is the mean squared error (MSE) loss. In the end, we sum up the outputs of the old trained model and the outputs of the BoostNet model for future inference. This method does not change the trained model but adds new weights, which come from the simpler model. The limitation is that the number of total weights increases every time a new dataset comes in. While in knowledge distillation (where a student model is trained to mimic the behavior of the teacher model), a base network was also trained and kept frozen~\cite{hinton2015distilling}, BoostNet aims to have different training targets  -- i.e., errors from former learners. 
 
\subsection{Frosting Network Weights by Meta-learning}
\label{Frosting Network Weights by Meta-learning}

We now explain our second ensemble model that can keep the number of parameters unchanged for inference. We add ``frosting'' layers for weights in a pre-trained model by taking weights as ``neurons.'' More specifically, we train a meta-network for weights in addition to the trained network. Both weights in the meta-network and weights in the base network are optimized during re-training. We have attempted freezing weights in the base network, but optimizing both networks yielded faster convergence. Each frosting layer is of the same shape as its corresponding weights. The weights in the frosting layers and the weights in the pre-trained model are both updated during retraining. In the end, the frosting layers are merged with the original weights so that the total number of weights is still unchanged. The merging operation is done by replacing the weights in the base network with the product of those weights and the weights in the corresponding frosting layer -- after which the frosting layers are discarded. The base network with the updated weights is used for inference -- thus, there is no performance loss after the merging operation. We call this ensemble model FrostNet, and it is illustrated in Figure~\ref{fig:FrostNet}. We have attempted common activation functions such as $tanh$ and $ReLU$; $tanh$ had a better performance compared to $ReLU$ in our experiments. Although $ReLU$ has many good metrics when applied to neurons, we do not find that it contributes to our frosting layer activation. Regardless, we have found that simply applying the frosting layers to trained weights without any activation functions already outperformed the benchmark models in our experiments.

\section{Experiments}

We now discuss the experiments used in evaluating the benefits of the proposed approaches.

\subsection{Datasets}
We use the MNIST and the CIFAR-10 datasets for comparing our BoostNet and FrostNet models with benchmark models. Similar to the data variation experiments in continual learning, we create a retraining dataset from the training set of the MNIST or CIFAR-10 by taking a random half of the original training set and augmenting the other half. The random half is taken by sampling each mini-batch during retraining instead of keeping them fixed before retraining. We obtain the augmented validation set using the same method. To this end, the sizes of the training and the validation sets stay unchanged in the initial training and retraining. We augment the entire test set as the new test set. The augmentations were conducted on the original training, validation, and test data separately to avoid data leakage. The augmentation factors are randomly rotating images by 45 degrees (clockwise and counterclockwise), shifting images by 20 percent (left and right), and zooming in by 80-90 percent. We have found that a well-trained model on the original MNIST or CIFAR-10 performs poorly on this augmentation setting.

Similarly to incremental class experiments in sequential learning, we split both the original MNIST and CIFAR-10 training datasets into two parts: the first part has classes 0-4; the second part has classes 5-9. We use the first part for training a LeNet model. We then take a random half of the first and the second parts during retraining. The original validation and test datasets are split into two parts under the same setting. Finally, we obtain a retraining dataset that is of the same size as the initial training dataset.

\begin{table}[t]
\centering
\begin{tabular}{|l|r|r|r|}
\hline
                    & Original & Augmented & Total   \\ \hline
Base                & 0.9921  & 0.2829  & 0.6352  \\ \hline        
Train-from-scratch  & 0.9870  & 0.7547  & 0.8699  \\ \hline
Fine-tuning         & 0.9921  & 0.2662  & 0.6267  \\ \hline
EWC                 & 0.9890  & 0.8117  & 0.8995  \\ \hline
MAS                 & 0.9889  & 0.6907  & 0.8385  \\ \hline
Selfless            & 0.9860  & 0.3036  & 0.6430  \\ \hline
BoostNet            & 0.9891  & 0.8830  & 0.9352  \\ \hline
FrostNet            & \textbf{0.9920}   & \textbf{0.9012}  & \textbf{0.9818} \\ \hline
\end{tabular}
\caption{Variational MNIST}
\label{MNIST-result}
\end{table}


\begin{table}[t]
\centering
\begin{tabular}{|l|r|r|r|}
\hline
                    & Original & Augmented & Total   \\ \hline
Base                & 0.6599  & 0.3013   & 0.4810  \\ \hline 
Train-from-scratch  & 0.6077  & 0.4309   & 0.5155  \\ \hline
Fine-tuning         & 0.6207  & 0.4463   & 0.5200  \\ \hline
EWC                 & 0.6191  & 0.4640   & 0.5415  \\ \hline
MAS                 & 0.6094  & 0.4363   & 0.5230  \\ \hline
Selfless            & 0.6198  & 0.3397   & 0.4795   \\ \hline
BoostNet            & 0.6363  & 0.4580   & 0.5426   \\ \hline
FrostNet            & \textbf{0.6456}  & \textbf{0.4646}   & \textbf{0.5552} \\ \hline
\end{tabular}
\caption{Variational CIFAR-10}
\label{cifar-10-result}
\end{table}

\begin{table}[t]
\centering
\begin{tabular}{|l|r|r|r|}
\hline
                    & Classes 0-4 & Classes 5-9 & Total   \\ \hline
Base                & 0.9984  & 0.0000  & 0.4992  \\ \hline        
Train-from-scratch  & 0.9957  & 0.9562  & 0.9760 \\ \hline
Fine-tuning         & 0.9953  & 0.9626  & 0.9790 \\ \hline
EWC                 & 0.9949  & 0.9562  & 0.9756  \\ \hline
MAS                 & 0.9916  & 0.9008  & 0.9462  \\ \hline
Selfless            & 0.9940  & 0.9074  & 0.9507  \\ \hline
BoostNet            & \textbf{0.9973}  & 0.9628  & 0.9800  \\ \hline
FrostNet            & 0.9963   & \textbf{0.9665}  & \textbf{0.9814} \\ \hline
\end{tabular}
\caption{Split MNIST}
\label{MNIST-Split-result}
\end{table}


\begin{table}[t]
\centering
\begin{tabular}{|l|r|r|r|}
\hline
                    & Classes 0-4 & Classes 5-9 & Total   \\ \hline
Base                & 0.7590  & 0.0000   & 0.3795  \\ \hline 
Train-from-scratch  & 0.4359  & \textbf{0.6518}   & 0.5439  \\ \hline
Fine-tuning         & 0.6010  & 0.5639   & 0.5824 \\ \hline
EWC                 & 0.5921  & 0.5520   & 0.5721  \\ \hline
MAS                 & 0.6402  & 0.5184   & 0.5230  \\ \hline
Selfless            & \textbf{0.6535}  & 0.4344  & 0.5440  \\ \hline
BoostNet            & 0.6252   & 0.5791  & 0.6022  \\ \hline
FrostNet            & 0.6498   & 0.5693  & \textbf{0.6096} \\ \hline
\end{tabular}
\caption{Split CIFAR-10}
\label{cifar-10-Split-result}
\end{table}

\subsection{Experimental Settings}
\label{Experimental Settings}
We compare our models to three baseline models: the initial LeNet model (base), training a new model (train-from-scratch), fine-tuning from pre-trained weights (fine-tuning), and three state-of-the-art models: EWC, MAS, and Selfless~\cite{aljundi2018selfless}. We use LeNet with batch normalization as the network framework. The pre-trained weights from the base model are used for measuring weight and neuron importance for EWC, MAS, and Selfless. All benchmark models are tuned based on the optimized weights from the base LeNet model. We use the truncated normal distribution for weight initialization for all network models. All models are trained at most 50 epochs with an early stopping of 10 consecutive epochs. The weights that have the highest accuracy and, secondarily, the lowest loss value (if the accuracy does not improve) on the validation dataset are used for inference. We report the accuracy of the corresponding test datasets in tables \ref{MNIST-result}--\ref{cifar-10-Split-result}. Boldface indicates the highest value in each column except the base model. We publish the code for our experiments on GitHub\footnote{https://github.com/XiaofengZhu/frosting\_weights}.

\section{Results}
\label{Results}

We average the accuracy of each model using three runs. Tables \ref{MNIST-result}-\ref{cifar-10-result} summarize the performance of the data variation experiments using the baseline models, the benchmark models, and our ensemble models. The base model has the highest accuracy on the original test datasets and the lowest accuracy on the augmented test datasets. Due to early-stopping and data augmentation, fine-tuning and benchmark models do not necessarily outperform train-from-scratch models. All the models except the base one suffer from catastrophic forgetting to some degree as the accuracy on an original test dataset is lower than the accuracy using the base model. All benchmark models have higher accuracy on the augmented dataset compared to the baseline models. EWC, MAS, and Selfless can prevent a trained model forgetting old knowledge, but their common limitation is that they have trouble learning from new knowledge. Our FrostNet model performs best among all benchmark models, and our BoostNet comes second. The FrostNet model yields the highest accuracy on the original datasets and the augmented datasets as the FrostNet model uses a network to learn how to adjust weights. The BoostNet model learns from performance gaps and pushes its weights to minimize the gaps.

Tables \ref{MNIST-Split-result}-\ref{cifar-10-Split-result} summarize the incremental class experiments. The base model has the highest accuracy on the datasets of classes (0-4) and the lowest accuracy on the ones of classes (5-9). Since the initial training dataset only has classes (0-4), the trained base model cannot predict any data from classes (5-9). Training-from-scratch on the retraining dataset has noticeable catastrophic forgetting on the initial training dataset. Although EWC, MAS, and especially Selfless models have an advantage in reducing catastrophic forgetting, they have inferior results on the new classes (5-9). Our BoostNet and FrostNet models can balance memorizing old knowledge and learning new knowledge, with a good overall performance.

\section{Conclusion}
\label{Conclusion}
In this paper, we propose two generic ensemble methods for boosting the performance of retraining. Regarding parameter capacity, BoostNet slowly adds a small number of parameters to a pre-trained model, similar to adding new trees in random forest modeling. Although FrostNet adds frosting layers for trained weights, the merging operation brings the number of total parameters down to the original size. Our work can be extended to other fields, such as general continual learning and network compression. Our work can be improved by taking the time series characteristic of training into account to further boost the performance. An interesting future direction would be strategically selecting training data. Another interesting direction would be selecting synapse paths from a sophisticated network model trained for a large domain to specialize for small domains.

\bibliographystyle{abbrv}
\bibliography{ref}

\begin{thebibliography}{10}

\bibitem{albesano2006adaptation}
D.~Albesano, R.~Gemello, P.~Laface, F.~Mana, and S.~Scanzio.
\newblock Adaptation of artificial neural networks avoiding catastrophic
  forgetting.
\newblock In {\em Proceedings of the 2006 IEEE International Joint Conference
  on Neural Network}, pages 1554--1561. IEEE, 2006.

\bibitem{aljundi2018memory}
R.~Aljundi, F.~Babiloni, M.~Elhoseiny, M.~Rohrbach, and T.~Tuytelaars.
\newblock Memory aware synapses: Learning what (not) to forget.
\newblock In {\em Proceedings of the European Conference on Computer Vision
  (ECCV)}, pages 139--154, 2018.

\bibitem{Aljundi2019}
R.~Aljundi, L.~Caccia, E.~Belilovsky, M.~Caccia, M.~Lin, L.~Charlin, and
  T.~Tuytelaars.
\newblock Online continual learning with maximally interfered retrieval.
\newblock 2019.

\bibitem{aljundi2018selfless}
R.~Aljundi, M.~Rohrbach, and T.~Tuytelaars.
\newblock Selfless sequential learning.
\newblock {\em arXiv preprint arXiv:1806.05421}, 2018.

\bibitem{chaudhry2018riemannian}
A.~Chaudhry, P.~K. Dokania, T.~Ajanthan, and P.~H. Torr.
\newblock Riemannian walk for incremental learning: Understanding forgetting
  and intransigence.
\newblock In {\em Proceedings of the European Conference on Computer Vision
  (ECCV)}, pages 532--547, 2018.

\bibitem{chaudhry2018efficient}
A.~Chaudhry, M.~Ranzato, M.~Rohrbach, and M.~Elhoseiny.
\newblock Efficient lifelong learning with a-gem.
\newblock {\em arXiv preprint arXiv:1812.00420}, 2018.

\bibitem{chaudhry2019continual}
A.~Chaudhry, M.~Rohrbach, M.~Elhoseiny, T.~Ajanthan, P.~K. Dokania, P.~H. Torr,
  and M.~Ranzato.
\newblock Continual learning with tiny episodic memories.
\newblock {\em arXiv preprint arXiv:1902.10486}, 2019.

\bibitem{chen2017learning}
G.~Chen, W.~Choi, X.~Yu, T.~Han, and M.~Chandraker.
\newblock Learning efficient object detection models with knowledge
  distillation.
\newblock In {\em Advances in Neural Information Processing Systems}, pages
  742--751, 2017.

\bibitem{Dai2017}
W.~Dai, Q.~Yang, G.-R. Xue, and Y.~Yu.
\newblock Boosting for transfer learning.
\newblock In {\em International Conference on Machine Learning}. IEEE, 2007.

\bibitem{Fernando2017}
C.~Fernando, D.~Banarse, C.~Blundell, Y.~Zwols, D.~Ha, A.~A. Rusu, A.~Pritzel,
  and D.~Wierstra.
\newblock Pathnet: Evolution channels gradient descent in super neural
  networks.
\newblock {\em CoRR}, abs/1701.08734, 2017.

\bibitem{french1999catastrophic}
R.~M. French.
\newblock Catastrophic forgetting in connectionist networks.
\newblock {\em Trends in cognitive sciences}, 3(4):128--135, 1999.

\bibitem{friedman2001greedy}
J.~H. Friedman.
\newblock Greedy function approximation: a gradient boosting machine.
\newblock {\em Annals of statistics}, pages 1189--1232, 2001.

\bibitem{Ganegedara2017}
T.~Ganegedara, L.~Ott, and F.~Ramos.
\newblock Lifelong learning with structurally adaptive cnns.
\newblock {\em International Conference on Machine Learning}, 2017.

\bibitem{henderson2017benchmark}
P.~Henderson, W.-D. Chang, F.~Shkurti, J.~Hansen, D.~Meger, and G.~Dudek.
\newblock Benchmark environments for multitask learning in continuous domains.
\newblock {\em International Conference on Machine Learning}, 2017.

\bibitem{hinton2015distilling}
G.~Hinton, O.~Vinyals, and J.~Dean.
\newblock Distilling the knowledge in a neural network.
\newblock {\em arXiv preprint arXiv:1503.02531}, 2015.

\bibitem{javed2019meta}
K.~Javed and M.~White.
\newblock Meta-learning representations for continual learning.
\newblock {\em arXiv preprint arXiv:1905.12588}, 2019.

\bibitem{juefei-xu2018pnn}
F.~Juefei-Xu, V.~N. Boddeti, and M.~Savvides.
\newblock {Perturbative Neural Networks}.
\newblock In {\em IEEE Computer Vision and Pattern Recognition (CVPR)}, June
  2018.

\bibitem{Nitin2017}
N.~Kamra, U.~Gupta, and Y.~Liu.
\newblock Deep generative dual memory network for continual learning.
\newblock abs/1710.10368, 2017.

\bibitem{kemker2017fearnet}
R.~Kemker and C.~Kanan.
\newblock Fearnet: Brain-inspired model for incremental learning.
\newblock {\em International Conference on Learning Representations (ICLR)},
  2017.

\bibitem{kirkpatrick2017overcoming}
J.~Kirkpatrick, R.~Pascanu, N.~Rabinowitz, J.~Veness, G.~Desjardins, A.~A.
  Rusu, K.~Milan, J.~Quan, T.~Ramalho, A.~Grabska-Barwinska, et~al.
\newblock Overcoming catastrophic forgetting in neural networks.
\newblock {\em Proceedings of the national academy of sciences},
  114(13):3521--3526, 2017.

\bibitem{kohonen1990self}
T.~Kohonen.
\newblock The self-organizing map.
\newblock {\em Proceedings of the IEEE}, 78(9):1464--1480, 1990.

\bibitem{lee2017overcoming}
S.-W. Lee, J.-H. Kim, J.~Jun, J.-W. Ha, and B.-T. Zhang.
\newblock Overcoming catastrophic forgetting by incremental moment matching.
\newblock In {\em Advances in neural information processing systems}, pages
  4652--4662, 2017.

\bibitem{li2018}
Y.~Li, Z.~Li, L.~Ding, P.~Yang, Y.~Hu, W.~Chen, and X.~Gao.
\newblock Supportnet: solving catastrophic forgetting in class incremental
  learning with support data.
\newblock {\em CoRR}, abs/1806.02942, 2018.

\bibitem{li2017learning}
Z.~Li and D.~Hoiem.
\newblock Learning without forgetting.
\newblock {\em IEEE transactions on pattern analysis and machine intelligence},
  40(12):2935--2947, 2017.

\bibitem{liang2019generative}
K.~J. Liang, C.~Li, G.~Wang, and L.~Carin.
\newblock Generative adversarial network training is a continual learning
  problem, 2019.

\bibitem{lin2013network}
M.~Lin, Q.~Chen, and S.~Yan.
\newblock Network in network.
\newblock {\em arXiv preprint arXiv:1312.4400}, 2013.

\bibitem{liu2018rotate}
X.~Liu, M.~Masana, L.~Herranz, J.~Van~de Weijer, A.~M. Lopez, and A.~D.
  Bagdanov.
\newblock Rotate your networks: Better weight consolidation and less
  catastrophic forgetting.
\newblock In {\em 2018 24th International Conference on Pattern Recognition
  (ICPR)}, pages 2262--2268. IEEE, 2018.

\bibitem{lopez2017gradient}
D.~Lopez-Paz et~al.
\newblock Gradient episodic memory for continual learning.
\newblock In {\em Advances in Neural Information Processing Systems}, pages
  6467--6476, 2017.

\bibitem{Luong2016}
Q.~V. L. I. S. O.~V. Luong, Minh-Thang and L.~Kaiser.
\newblock Multi-task sequence to sequence learning.
\newblock {\em International Conference on Learning Representations (ICLR)},
  2016.

\bibitem{mallya2018packnet}
A.~Mallya and S.~Lazebnik.
\newblock Packnet: Adding multiple tasks to a single network by iterative
  pruning.
\newblock In {\em Proceedings of the IEEE Conference on Computer Vision and
  Pattern Recognition}, pages 7765--7773, 2018.

\bibitem{Maltoni2018}
D.~Maltoni and V.~Lomonaco.
\newblock Continuous learning in single-incremental-task scenarios.
\newblock {\em CoRR}, abs/1806.08568, 2018.

\bibitem{Mehta2018}
S.~V. Mehta, B.~Paranjape, and S.~Singh.
\newblock Evaluating influence functions for memory replay in continual
  learning.
\newblock 2019.

\bibitem{Meyerson2018}
E.~Meyerson and R.~Miikkulainen.
\newblock Beyond shared hierarchies: Deep multitask learning through soft layer
  ordering.
\newblock {\em International Conference on Learning Representations (ICLR)},
  2018.

\bibitem{Elliot2018}
E.~Meyerson and R.~Miikkulainen.
\newblock Pseudo-task augmentation: From deep multitask learning to intratask
  sharing - and back.
\newblock 2018.

\bibitem{nguyen2018variational}
C.~V. Nguyen, Y.~Li, T.~D. Bui, and R.~E. Turner.
\newblock Variational continual learning.
\newblock In {\em International Conference on Learning Representations}, 2018.

\bibitem{DBLP:journals/corr/abs-1802-07569}
G.~I. Parisi, R.~Kemker, J.~L. Part, C.~Kanan, and S.~Wermter.
\newblock Continual lifelong learning with neural networks: {A} review.
\newblock {\em CoRR}, abs/1802.07569, 2018.

\bibitem{polikar2001learn++}
R.~Polikar, L.~Upda, S.~S. Upda, and V.~Honavar.
\newblock Learn++: An incremental learning algorithm for supervised neural
  networks.
\newblock {\em IEEE transactions on systems, man, and cybernetics, part C
  (applications and reviews)}, 31(4):497--508, 2001.

\bibitem{rebuffi2017icarl}
S.-A. Rebuffi, A.~Kolesnikov, G.~Sperl, and C.~H. Lampert.
\newblock icarl: Incremental classifier and representation learning.
\newblock In {\em Proceedings of the IEEE Conference on Computer Vision and
  Pattern Recognition}, pages 2001--2010, 2017.

\bibitem{Ren2017}
H.~W. J.~L. Ren, Boya and H.~Gao.
\newblock Life-long learning based on dynamic combination model.
\newblock 2017.

\bibitem{Rios2018}
A.~Rios and L.~Itti.
\newblock Closed-loop {GAN} for continual learning.
\newblock {\em CoRR}, abs/1811.01146, 2018.

\bibitem{roy2018tree}
D.~Roy, P.~Panda, and K.~Roy.
\newblock Tree-cnn: a hierarchical deep convolutional neural network for
  incremental learning.
\newblock {\em arXiv preprint arXiv:1802.05800}, 2018.

\bibitem{Ruder2017}
S.~Ruder.
\newblock An overview of multi-task learning in deep neural networks.
\newblock {\em CoRR}, abs/1706.05098, 2017.

\bibitem{rusu2016progressive}
A.~A. Rusu, N.~C. Rabinowitz, G.~Desjardins, H.~Soyer, J.~Kirkpatrick,
  K.~Kavukcuoglu, R.~Pascanu, and R.~Hadsell.
\newblock Progressive neural networks.
\newblock {\em arXiv preprint arXiv:1606.04671}, 2016.

\bibitem{Rusu2016}
A.~A. Rusu, N.~C. Rabinowitz, G.~Desjardins, H.~Soyer, J.~Kirkpatrick,
  K.~Kavukcuoglu, R.~Pascanu, and R.~Hadsell.
\newblock Progressive neural networks.
\newblock {\em CoRR}, abs/1606.04671, 2016.

\bibitem{ruvolo2013ella}
P.~Ruvolo and E.~Eaton.
\newblock Ella: An efficient lifelong learning algorithm.
\newblock In {\em International Conference on Machine Learning}, pages
  507--515, 2013.

\bibitem{Schwarz2018}
J.~L. W. M. C. A. G.-B. Y. W. T. R.~P. Schwarz, Jonathan and R.~Hadsell.
\newblock Progress \& compress: A scalable framework for continual learning.
\newblock 2018.

\bibitem{Seff2017}
A.~Seff, A.~Beatson, D.~Suo, and H.~Liu.
\newblock Continual learning in generative adversarial nets.
\newblock {\em CoRR}, abs/1705.08395, 2017.

\bibitem{serra2018overcoming}
J.~Serr{\`a}, D.~Sur{\'\i}s, M.~Miron, and A.~Karatzoglou.
\newblock Overcoming catastrophic forgetting with hard attention to the task.
\newblock {\em arXiv preprint arXiv:1801.01423}, 2018.

\bibitem{shin2017continual}
H.~Shin, J.~K. Lee, J.~Kim, and J.~Kim.
\newblock Continual learning with deep generative replay.
\newblock In {\em Advances in Neural Information Processing Systems}, pages
  2990--2999, 2017.

\bibitem{silver2013lifelong}
D.~L. Silver, Q.~Yang, and L.~Li.
\newblock Lifelong machine learning systems: Beyond learning algorithms.
\newblock In {\em 2013 AAAI spring symposium series}, 2013.

\bibitem{Sodhani2019}
S.~Sodhani, S.~Chandar, and Y.~Bengio.
\newblock On training recurrent neural networks for lifelong learning.
\newblock {\em International Conference on Machine Learning}, 2019.

\bibitem{standley2019tasks}
T.~Standley, A.~R. Zamir, D.~Chen, L.~Guibas, J.~Malik, and S.~Savarese.
\newblock Which tasks should be learned together in multi-task learning?
\newblock {\em International Conference on Machine Learning}, 2019.

\bibitem{swaroop2019improving}
S.~Swaroop, C.~V. Nguyen, T.~D. Bui, and R.~E. Turner.
\newblock Improving and understanding variational continual learning.
\newblock {\em arXiv:1905.02099 [cs, stat]}, 2019.

\bibitem{Subakan2019}
C.~Sübakan, M.~Caccia, T.~Lesort, and L.~Charlin.
\newblock Continual learning of generative models with maximum entropy
  generative replay.
\newblock 2019.

\bibitem{Thanh-Tung2018}
H.~Thanh{-}Tung, T.~Tran, and S.~Venkatesh.
\newblock On catastrophic forgetting and mode collapse in generative
  adversarial networks.
\newblock {\em CoRR}, abs/1807.04015, 2018.

\bibitem{tu2016ranking}
M.~Tu, V.~Berisha, M.~Woolf, J.-s. Seo, and Y.~Cao.
\newblock Ranking the parameters of deep neural networks using the fisher
  information.
\newblock In {\em Proceedings of 2016 IEEE International Conference on
  Acoustics, Speech and Signal Processing (ICASSP)}, pages 2647--2651. IEEE,
  2016.

\bibitem{wang2017model}
C.~Wang, X.~Lan, and Y.~Zhang.
\newblock Model distillation with knowledge transfer from face classification
  to alignment and verification.
\newblock {\em arXiv preprint arXiv:1709.02929}, 2017.

\bibitem{xu2018reinforced}
J.~Xu and Z.~Zhu.
\newblock Reinforced continual learning.
\newblock In {\em Advances in Neural Information Processing Systems}, pages
  899--908, 2018.

\bibitem{yoon2018lifelong}
J.~Yoon, E.~Yang, J.~Lee, and S.~J. Hwang.
\newblock Lifelong learning with dynamically expandable networks.
\newblock {\em International Conference on Learning Representations (ICLR)},
  2018.

\bibitem{yosinski2014transferable}
J.~Yosinski, J.~Clune, Y.~Bengio, and H.~Lipson.
\newblock How transferable are features in deep neural networks?
\newblock In {\em Advances in neural information processing systems}, pages
  3320--3328, 2014.

\bibitem{YuSlimmable}
J.~Yu, L.~Yang, N.~Xu, J.~Yang, and T.~Huang.
\newblock Slimmable neural networks.
\newblock {\em International Conference on Learning Representations (ICLR)},
  2019.

\bibitem{zenke2017continual}
F.~Zenke, B.~Poole, and S.~Ganguli.
\newblock Continual learning through synaptic intelligence.
\newblock In {\em Proceedings of the 34th International Conference on Machine
  Learning-Volume 70}, pages 3987--3995. JMLR. org, 2017.

\bibitem{zeno2018task}
C.~Zeno, I.~Golan, E.~Hoffer, and D.~Soudry.
\newblock Task agnostic continual learning using online variational bayes.
\newblock {\em arXiv preprint arXiv:1803.10123}, 2018.

\bibitem{zhang2017}
Y.~Zhang and Q.~Yang.
\newblock A survey on multi-task learning.
\newblock {\em CoRR}, abs/1707.08114, 2017.

\end{thebibliography}

\end{document}